%% file: main.tex
\documentclass{article}
\usepackage{hyphenat}

\usepackage[preprint]{neurips_2025}

\usepackage[utf8]{inputenc} 
\usepackage[T1]{fontenc}    
\usepackage{hyperref}       
\usepackage{url}            
\usepackage{booktabs}       
\usepackage{amsfonts}       
\usepackage{nicefrac}       
\usepackage{microtype}      
\usepackage{xcolor}         
\usepackage{graphicx}
\usepackage[noframe]{showframe}
\usepackage{framed}
\usepackage{lipsum}
\usepackage{enumitem}
\usepackage{booktabs}
\usepackage{amsmath}
\renewenvironment{shaded}{%
  \MakeFramed{\advance\hsize-\width \FrameRestore\FrameRestore}}%
 {\endMakeFramed}
\definecolor{shadecolor}{gray}{0.75}

\title{SmartAvatar: Text- and Image-Guided Human Avatar Generation with VLM AI Agents}

\author{%
  \begin{tabular}{c}  
    Alexander~Huang\textendash Menders$^{2}$,
    Xinhang~Liu$^{1}$,
    Andy~Xu$^{2}$,
    Yuyao~Zhang$^{2}$\\
    Chi\textendash Keung~Tang$^{1}$,
    Yu\textendash Wing~Tai$^{2}$
  \end{tabular}\\[1em]%
  $^{1}$The Hong Kong University of Science and Technology\qquad
  $^{2}$Dartmouth College%
}
\begin{document}

\maketitle

\input{sections/abstract}
\input{sections/intro}
\input{sections/related_works}

\input{sections/methods}

\input{sections/experiments}
\input{sections/ablation_study}

\input{sections/conclusion}

{
\small
\bibliographystyle{plainnat}
\bibliography{main}
}

\input{sections/supp}
\end{document}

%% file: sections/abstract.tex
\begin{abstract}
SmartAvatar is a vision-language-agent-driven framework for generating fully rigged, animation-ready 3D human avatars from a single photo or textual prompt. While diffusion-based methods have made progress in general 3D object generation, they continue to struggle with precise control over human identity, body shape, and animation readiness. In contrast, SmartAvatar leverages the commonsense reasoning capabilities of large vision-language models (VLMs) in combination with off-the-shelf parametric human generators to deliver high-quality, customizable avatars. A key innovation is an autonomous verification loop, where the agent renders draft avatars, evaluates facial similarity, anatomical plausibility, and prompt alignment, and iteratively adjusts generation parameters for convergence. This interactive, AI-guided refinement process promotes fine-grained control over both facial and body features, enabling users to iteratively refine their avatars via natural-language conversations. Unlike diffusion models that rely on static pre-trained datasets and offer limited flexibility, SmartAvatar brings users into the modeling loop and ensures continuous improvement through an LLM-driven procedural generation and verification system. The generated avatars are fully rigged and support pose manipulation with consistent identity and appearance, making them suitable for downstream animation and interactive applications. Quantitative benchmarks and user studies demonstrate that SmartAvatar outperforms recent text- and image-driven avatar generation systems in terms of reconstructed mesh quality, identity fidelity, attribute accuracy, and animation readiness, making it a versatile tool for realistic, customizable avatar creation on consumer-grade hardware.
\end{abstract}

%% file: sections/intro.tex
\section{Introduction}

Creating high-quality, controllable, and animatable 3D human avatars is essential for applications in gaming, virtual reality, digital fashion, and telepresence. Despite substantial progress in generative AI across both 2D~\cite{rombach2022high, zhang2023adding} and 3D domains~\cite{poole2022dreamfusion, lin2023magic3d, hong2023lrm}, producing realistic, riggable human avatars that are faithful to user input and ready for animation remains an open challenge. Diffusion-based approaches can synthesize visually rich avatars from textual or visual prompts~\cite{zhang2024clay, siddiqui2024meta}, but they struggle with precise control over identity and pose, often demand significant computational resources, and typically lack support for animation or customization.

Parametric human generation methods, such as SMPL~\cite{smpl,smpl-x}, AG3D\cite{ag3d}, AvatarGen\cite{avatargen}, EVA3D\cite{eva3d} and HumanGen3D~\cite{humgen3d}, offer a more tractable alternative. These systems support real-time, riggable 3D avatars with explicit control over key features such as body shape, facial geometry, clothing, and pose. As a result, they are widely adopted in production pipelines where consistency, editability, and motion compatibility are crucial. However, leveraging these systems at scale remains difficult. 

\begin{figure}[t]
\vspace{-0.1in}
    \centering
    \includegraphics[width=1\linewidth]{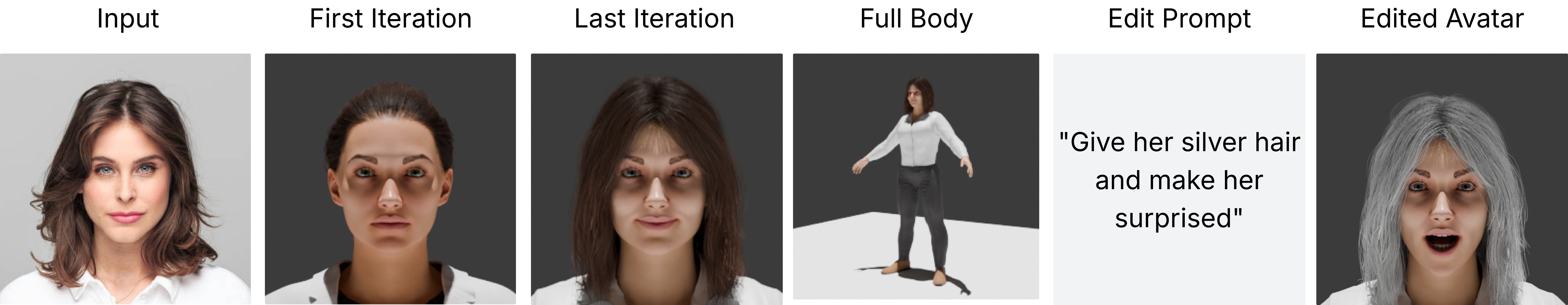}
    \small\textit{(a) Generation and editing results for an image-only portrait input.}
 
    \includegraphics[width=1\linewidth]{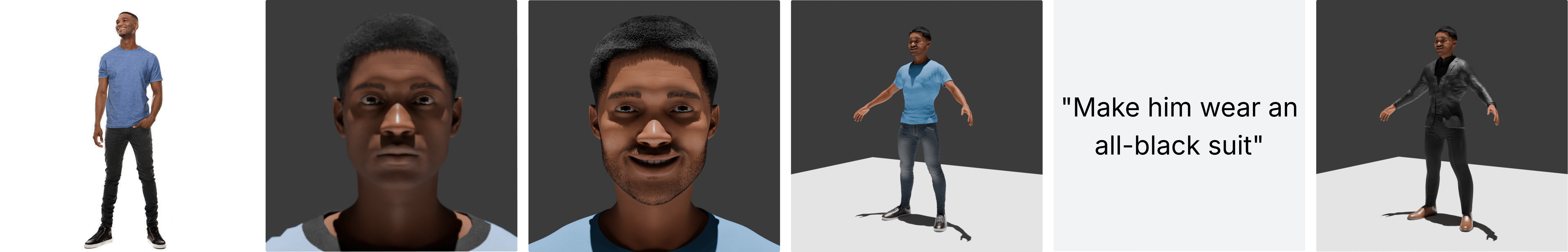}
    \small\textit{(b) Generation and editing results for an image-only full body input.}

    \includegraphics[width=1\linewidth]{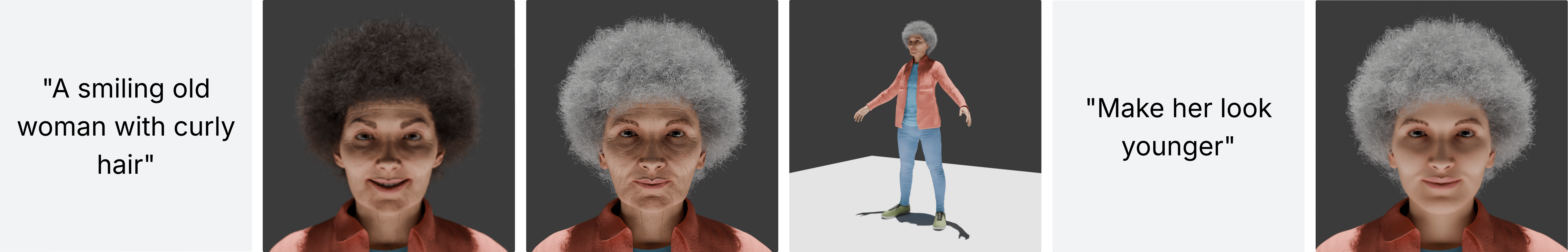}
    \small\textit{(c) Generation and editing results for an text-only input.}

    \includegraphics[width=1\linewidth]{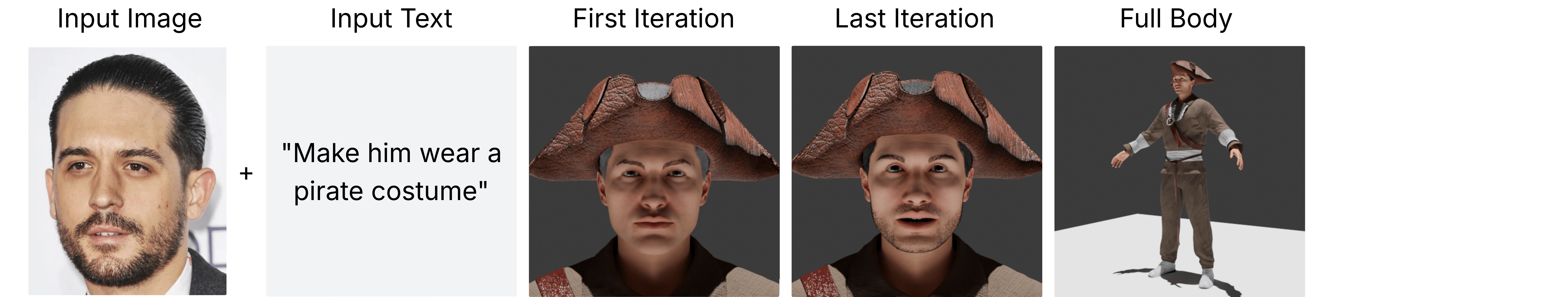}
    \small\textit{(d) Generation results for a joint image and text input.}
    \vspace{-0.05in}
    \caption{\textbf{Examples of SmartAvatar’s iterative generation and editing pipeline.} The system takes four types of input: (a) portrait image, (b) full-body image, (c) text-only description, and (d) a combination of image and text. For each input type, SmartAvatar generates an initial draft avatar and then refines it through auto-verification loops guided by a vision-language agent. The figure shows both the initial and final outputs, including portrait and full-body views. In cases (a) through (c), we also illustrate the editing capabilities of the system, where avatars can be modified further using natural language prompts. These results highlight SmartAvatar’s ability to produce photorealistic, animation-ready avatars with strong identity and attribute consistency from a wide range of input modalities.} \vspace{-0.25in}
    \label{fig:teaser}
\end{figure}

Translating natural language or image inputs into detailed generator parameters is inherently complex, typically requiring expert knowledge or manually constructed mappings. Moreover, most current pipelines lack mechanisms to verify that the output accurately reflects the user's intent.

To overcome these limitations, we introduce SmartAvatar, a vision-language-agent-driven framework for generating fully rigged, photorealistic, and animation-ready 3D human avatars from a single image or text prompt. At the core of SmartAvatar is a GPT-4o-style vision-language model (VLM) agent~\cite{achiam2023gpt} that interprets user inputs, operates parametric avatar generators, and allows conversational refinement using natural language. This makes SmartAvatar accessible to non-experts while enabling detailed, iterative control for advanced users.

A central innovation in our approach is an auto-verification loop, where the VLM agent renders interim avatar drafts and evaluates them against the input using multiple criteria: facial similarity, anatomical plausibility, attribute alignment, and perceptual coherence. This dynamic refinement mechanism allows SmartAvatar to continuously adjust generator parameters until the output closely matches the user’s intent, bridging the gap between symbolic control and perceptual fidelity. 

Extensive qualitative evaluations and visual comparisons show that SmartAvatar surpasses existing state-of-the-art text-to-human and image-to-human methods in terms of reconstructed mesh quality, attribute fidelity, user-specific customization, and animation readiness, all while maintaining efficient performance on consumer-grade hardware.

Our key contributions are as follows:

\begin{itemize}[leftmargin=5mm]
    \item We present SmartAvatar, a vision-language-agent-driven framework for generating customizable, rigged 3D human avatars from a single image or text prompt.
    \item We introduce a VLM-guided auto-verification loop that evaluates and iteratively refines generated avatars to match user input across visual and semantic criteria.
    \item We enable conversational avatar editing via natural-language interactions, dramatically lowering the barrier to high-quality 3D avatar creation.
    \item We demonstrate through qualitative benchmarks and visual comparisons that SmartAvatar outperforms existing methods in reconstructed mesh quality, attribute control, and animation readiness.
\end{itemize}

%% file: sections/related_works.tex
\section{Related Works}

\noindent\textbf{Parametric Models for 3D Human Generation.}
Parametric human models such as SMPL~\cite{smpl} and its extension SMPL-X~\cite{smpl-x} have become foundational tools in avatar generation, offering efficient and anatomically plausible representations of human body shape and pose. These models have enabled a broad range of follow-up works in 3D human generation, such as AG3D~\cite{ag3d}, AvatarGen~\cite{avatargen}, and EVA3D~\cite{eva3d} to synthesize 3D human avatars with relatively high fidelity and controllability. However, their linear structure imposes inherent limitations in capturing fine-grained facial details and non-rigid body deformations. This restricts their expressiveness in interactive applications where realism, identity fidelity, and nuanced pose articulation are critical~\cite{avatargen, smpl-x}.

\vspace{1mm}
\noindent\textbf{Diffusion-Based 3D Modeling.}
Diffusion models have made significant strides in producing high-quality 2D and 3D content~\cite{xiang2025structured3dlatentsscalable, li2025pshumanphotorealisticsingleimage3d, ramesh2021zeroshottexttoimagegeneration, saharia2022photorealistictexttoimagediffusionmodels}. Structured latent diffusion models such as StructLDM~\cite{structldm} allow localized edits by organizing content into semantically structured latent spaces. In the context of human avatar generation, DreamFusion-style pipelines~\cite{kolotouros2023dreamhuman, jiang2023avatarcraft, NEURIPS2023_0e769ec2} use Score Distillation Sampling (SDS) to guide 3D optimization using 2D diffusion priors. Variants like DreamHuman~\cite{kolotouros2023dreamhuman} and DreamWaltz~\cite{NEURIPS2023_0e769ec2} add pose-conditioning or skeleton constraints to improve structure. Other techniques ~\cite{Liu_2024_CVPR} extend to 3D gaussian splatting to render human structures. Despite these advancements, diffusion-based approaches remain limited by long generation times, lack of animation readiness, and poor control over identity, geometry, and semantic attributes. They are also prone to common artifacts such as the Janus effect and content drift, especially in occluded regions.

\vspace{1mm}
\noindent\textbf{LLM-Driven Agentic Generation.}
The rise of large language models (LLMs) and multimodal vision-language models (VLMs) has enabled generative agents capable of planning, reasoning, and interacting across complex environments~\cite{achiam2023gpt, team2023gemini, alayrac2022flamingo, liu2023visual, zhu2023minigpt, liu2023blip2}. These models support a new paradigm of interactive content generation, moving beyond static one-shot methods. Agentic systems~\cite{park2023generativeagents, jiang2023voyager, liu2023agentbench, yang2024worldgpt, hu2024scenecraft, wu2024motion} demonstrate how VLMs can synthesize coherent 3D scenes, avatars, and behaviors by integrating natural language understanding with procedural generation. Crucially, recent frameworks employ self-refinement loops in which the model serves as both generator and evaluator~\cite{madaan2023selfrefineiterativerefinementselffeedback, radha2024iterationthoughtleveraginginner, ashutosh2025llmsheartraining}, iteratively critiquing and updating outputs to improve quality without retraining.

\vspace{1mm}
\noindent\textbf{Our Contribution.}
SmartAvatar builds upon these prior strands of research but offers a unified framework that overcomes their key limitations. Unlike diffusion pipelines that rely on static, slow, and often uncontrollable optimization, SmartAvatar uses a VLM-powered agent to guide off-the-shelf parametric human models~\cite{humgen3d} through an iterative, feedback-driven process. This includes rendering candidate avatars, evaluating them for identity fidelity, anatomical plausibility, and prompt alignment, and autonomously refining the model until convergence. By placing a reasoning-capable agent in the loop, SmartAvatar enables nuanced avatar customization from text or images, robust facial and body feature control, and seamless integration with motion synthesis for animation. This VLM-guided, self-verifying pipeline transforms avatar creation from a passive generation task into an interactive design loop, supporting real-time, high-quality avatar production on consumer-grade hardware.

%% file: sections/methods.tex
\section{Method}
\label{sec:method}

\begin{figure}[ht]
\centering
\includegraphics[width=\linewidth]{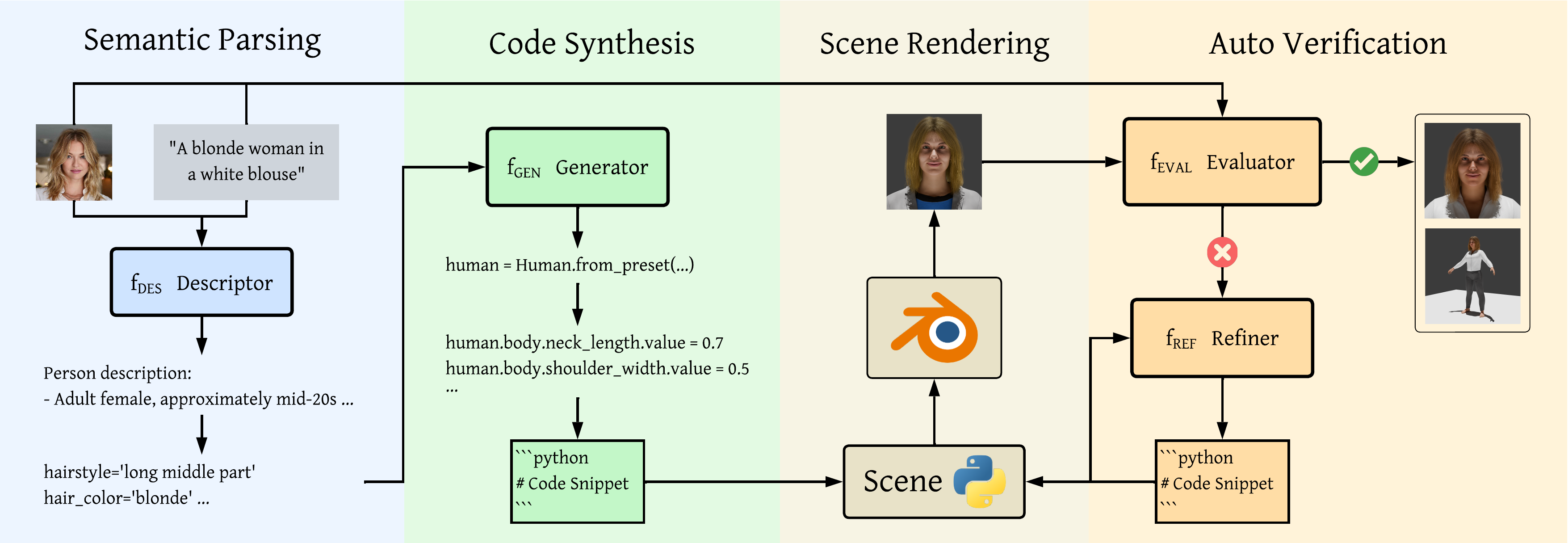}
\caption{SmartAvatar pipeline: (1) Descriptor LLM extracts semantic attributes; (2) Generator LLM creates Blender Python code; (3) Refiner LLM optionally adjusts code based on visual feedback; (4) final avatar is rendered when similarity exceeds a threshold or max iterations are reached.}\vspace{-0.1in}
\label{fig:pipeline}
\end{figure}

\noindent\textbf{Overview.}
SmartAvatar is a vision-language-agent-driven system for generating fully rigged, animation-ready 3D human avatars from either a single image or a text description. It orchestrates four large language models (LLMs) in a modular agent pipeline (Figure~\ref{fig:pipeline}): a \textit{Descriptor} to extract semantic attributes, a \textit{Generator} to produce Blender-compatible Python code, an \textit{Evaluator} to compare the rendered avatar with the input and assess identity and attribute alignment, and a \textit{Refiner} to iteratively improve the avatar based on the Evaluator’s feedback. The system leverages chain-of-thought (CoT) reasoning and a verification loop to ensure fidelity, editability, and compatibility with animation workflows.

\noindent\textbf{Input-to-Avatar Pipeline.}
Given an input—either an image  $I_{\mathrm{orig}}$, or a text description $T$—SmartAvatar proceeds through four stages:

\begin{enumerate}[leftmargin=5mm]
    \item \textbf{Semantic Parsing:} The descriptor LLM $f_{\mathrm{des}}$ is responsible for extracting a structured attribute representation from multimodal input—either a portrait or full-body image $I_{\mathrm{orig}}$, a natural-language description $T$, or both. It operates in conjunction with a predefined avatar construction API manual~\cite{humgen3d}, which enumerates all valid modifiable attributes (e.g., face shape, hairstyle, attire types) and acceptable parameter values.
\begin{itemize}[leftmargin=5mm]
\item \textbf{Text Input (CoT Reasoning):} For purely textual prompts, $f_{\mathrm{des}}$—powered by GPT-4o—employs chain-of-thought (CoT) reasoning to decompose the high-level description into a set of explicit avatar properties. For example, given a prompt like \textit{“A basketball player”}, the agent infers attributes such as:
\begin{itemize}
\item Occupation $\rightarrow$ athletic build, muscular frame
\item Context $\rightarrow$ basketball uniform, high-top sneakers
\item Age/gender prior $\rightarrow$ young adult male (if unspecified)
\end{itemize}
These decisions are serialized as key-value tokens, aligned to the API specification, and used to drive downstream generation.
\item \textbf{Image Input (VLM Extraction):} For visual prompts, SmartAvatar uses GPT-4o’s multimodal capabilities to extract facial and stylistic attributes. The image is passed through the model with CoT-style prompts that encourage explicit reasoning about visible features, such as skin tone, facial hair, hairstyle, clothing type, and color palette.

\item \textbf{Multimodal Fusion:} When both image and text inputs are provided, the agent jointly reasons over the two modalities. Attributes from the image serve as anchors, while textual descriptions provide overriding intent or contextual clues. A priority heuristic is used to resolve conflicts—for example, if the image shows long hair but the text says “shaved head,” the text is assumed to be a desired modification.

\item \textbf{API-Aware Normalization:} After attribute extraction, the LLM maps free-form descriptions into discrete, API-compatible tokens. For instance, “messy brown hair” is parsed into \texttt{hairstyle=‘messy’}, \texttt{hair\_color=‘brown’}. The API manual—loaded into the LLM’s context—helps constrain this reasoning, ensuring only valid categories are emitted.
\end{itemize}
    \item \textbf{Code Synthesis:} The generator LLM $f_{\mathrm{gen}}$ receives the structured attribute representation from the descriptor and performs chain-of-thought reasoning to translate it into Blender-compatible Python code. The synthesis process follows a multi-step planning sequence:

\begin{enumerate}[leftmargin=5mm]
\item \textbf{Morph Base Selection:} The generator selects a suitable base mesh variant conditioned on body type, gender presentation, and age attributes.
\item \textbf{Component Attachment:} The model sequentially applies modifiers to the face, hair, clothing, and accessories. Each step maps semantic tokens (e.g., "ponytail", "glasses", "formal shirt") to predefined asset libraries or parametric code primitives within a constrained Blender API.
\item \textbf{Code Emission:} The LLM emits executable Python code conforming to a strict schema, including function calls for model import, mesh deformation, material assignment, and rig binding.
\end{enumerate}

To support reproducibility and prevent regressions, SmartAvatar maintains a growing repository of verified code examples. After each synthesis pass:

\begin{itemize}[leftmargin=5mm]
\item If code execution fails (e.g., due to syntax errors, missing asset references, or API misuse), the error trace is captured and forwarded back to $f_{\mathrm{gen}}$ for diagnosis. The model reasons about the failure and emits a corrected version of the code.
\item Successfully executed scripts are automatically validated and appended to the code example bank as canonical references. These examples serve two purposes: (1) guiding future code generation through retrieval-augmented prompting, and (2) informing few-shot demonstrations to reduce hallucination in subsequent synthesis steps.
\end{itemize}

All generated code is sandboxed and subjected to static analysis and type checking before execution. Only code that passes both syntactic and functional validation proceeds to the rendering stage.
    \item \textbf{Scene Rendering:} The code is embedded in a template with fixed scene parameters (camera, lighting) and rendered in Blender to produce an initial avatar image $I_{\mathrm{rend}}$.
    \item \textbf{Auto-Verification Loop:} An evaluator module determines if the rendered avatar is similar enough to the input to determine if the output should be iteratively refined further or not. This task can be offloaded to an evaluator VLM, $f_{\mathrm{eval}}$, for both image and/or text input. Alternatively, for image inputs, a visual encoder $F$ can compute cosine similarity between the original input and the rendered avatar:
\begin{equation}
s = \cos(F(I_{\mathrm{orig}}), F(I_{\mathrm{rend}}))
\end{equation}
If $s < \tau$ and iteration $i < N_{\max}$, the refiner LLM $f_{\mathrm{ref}}$ is invoked to revise the code. For portrait-style inputs, similarity is computed primarily over facial embeddings. However, if the input is a full-body image, the system additionally evaluates hair style, clothing color, and garment type consistency using a vision-language model (VLM). The VLM generates a set of revision suggestions in the form of updated attribute tokens and/or parameterized semantic descriptions. These serve as conditioning input for the refiner LLM $f_{\mathrm{ref}}$, which re-synthesizes Blender code accordingly. The process iterates until the similarity score meets the threshold ($s \geq \tau$) or the maximum number of iterations is reached ($i = N_{\max}$).
\end{enumerate}

\noindent\textbf{Chain-of-Thought and Agent Communication.}
Each LLM in the pipeline follows a structured CoT prompting strategy:
\begin{itemize}[leftmargin=5mm]
\item The descriptor parses high-level descriptions into discrete attributes.
\item The generator follows a reasoning chain: select morph $\rightarrow$ apply edits $\rightarrow$ assign clothing $\rightarrow$ finalize materials.
\item Inter-agent communication occurs via structured messages containing prompt context, semantic attributes, image feedback, and Blender API logs.
\end{itemize}

This modular reasoning reduces hallucination and improves consistency. Intermediate decisions are serialized into JSON-like logs for traceability across stages.

\noindent\textbf{Refinement Loop.}  
When the similarity score falls below a threshold, the evaluator function \( f_{\mathrm{eval}} \) and the refiner \( f_{\mathrm{ref}} \) operate together as a feedback-aware controller:
\begin{enumerate}[leftmargin=5mm]
    \item They take as input the previous generation code, the rendered output, and the similarity score.
    \item They analyze discrepancies between the input and output, identifying mismatches such as incorrect skin tone or hairstyle.
    \item They generate delta code or parameter updates to correct the avatar, which is then re-rendered for further evaluation.
\end{enumerate}

\noindent\textbf{User Customization.}
After generation, users can issue free-form textual instructions (e.g., \textit{“change to military uniform”}). These are routed to the refinement loop, which performs constrained edits while preserving identity and scene integrity. Samples in Figure~\ref{fig:teaser} demonstrate the editing results after generating a new avatar.

\noindent\textbf{Multimodal Unification.}
All inputs, whether images or text, are assigned to a shared semantic attribute space to unify downstream processing. This ensures consistent behavior across input types.
All code is validated before execution using static type checks and a strict API schema to ensure safety and reproducibility.

\noindent\textbf{Human Generator.}
We leverage the open-source Human Generator 3D (HumGen3D) add-on for Blender as the backbone of our avatar construction pipeline~\cite{humgen3d}. HumGen3D provides a flexible API for programmatically modifying photorealistic, fully rigged human models, including attributes such as facial structure, skin tone, hairstyle, clothing, and body shape. Its parametric control interface enables precise mesh deformation and rigging through Python scripting, making it well-suited for LLM-driven code generation. All avatar generation within SmartAvatar occurs within a headless Blender environment using the HumGen3D model, ensuring consistent base geometry, animation compatibility, and material realism. We load the HumGen3D API specification into the Descriptor LLM’s context, allowing the agent to constrain outputs to valid modifiable attributes and generate Blender-compatible function calls. This setup provides a high-fidelity, scriptable avatar workspace that supports diverse identities and customization options.

\noindent\textbf{Discussion.}
The modular agent-based pipeline in SmartAvatar offers several advantages over monolithic generation approaches. By separating semantic parsing, code synthesis, and feedback-based refinement, each LLM operates within a clearly defined scope, enhancing interpretability, traceability, and controllability. The chain-of-thought strategy ensures that reasoning steps remain transparent and debuggable, while the auto-verification loop provides a principled way to align output with the input identity. Moreover, unifying multimodal inputs into a shared attribute space facilitates consistent avatar generation across both visual and textual modalities. This design not only improves fidelity and editability but also allows for scalable extensions, such as plug-in motion modules or domain-specific avatar templates.

%% file: sections/experiments.tex
\section{Experimental Results}
We evaluate SmartAvatar on a suite of scenarios to validate its ability to generate high-quality, animation-ready 3D human avatars from both image and text inputs. The experiments demonstrate the framework's strengths in multimodal control, editability, and high-fidelity reconstruction, while also highlighting its robust agent pipeline for iterative refinement.

\begin{figure}
    \centering
    \includegraphics[width=1\linewidth]{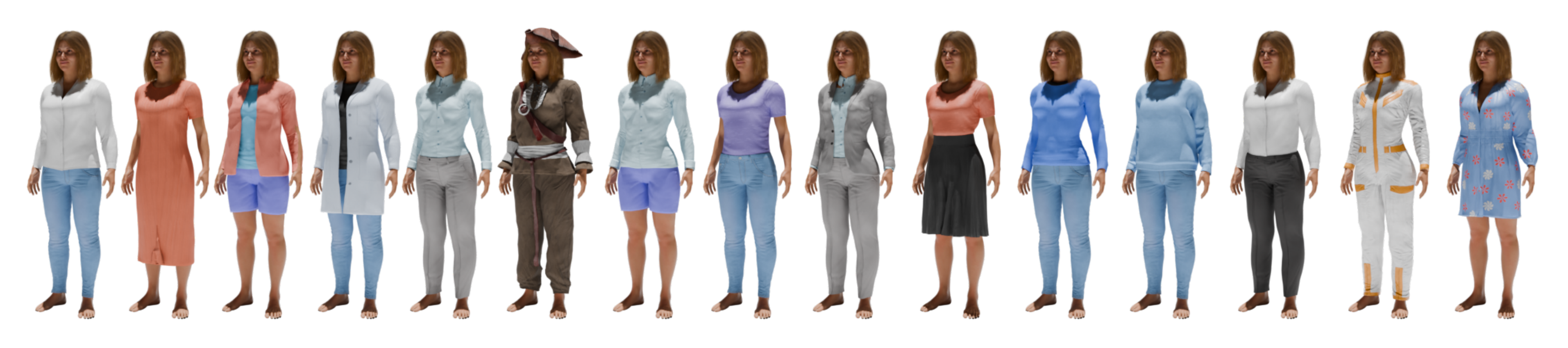}
    \vspace{-0.3in}
    \caption{Snapshot of subset of outfit diversity enabled by the human generator. Our pipeline selects from a wide array of outfit presets and can go further to manipulate color of clothing. In this example, we generated an avatar using the SmartAvatar pipeline and then randomized outfits to display clothing diversity on the same avatar.}
    \label{fig:outfits}
    \centering
    \includegraphics[width=1\linewidth]{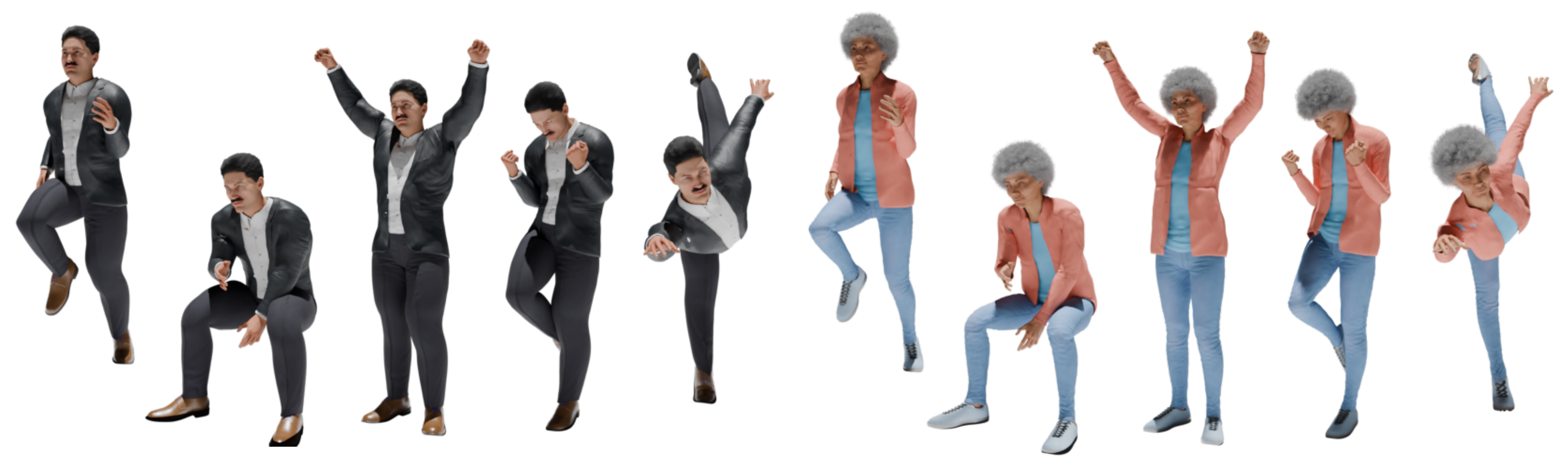}
    \vspace{-0.3in}
    \caption{Pose Manipulation Examples: The generated avatars are fully rigged and can be posed using a wide range of preset or custom poses for animation. Their appearance remains consistent across different poses.}
    \vspace{-0.2in}
    \label{fig:poses}
\end{figure}

\noindent\textbf{Diverse Avatar Generation}
Figure~\ref{fig:outfits} showcases SmartAvatar's generation and editing capabilities, specifically demonstrating outfit modifications on a generated avatar. While this example focuses on clothing edits, our method supports a wide variety of outputs driven by natural language prompts. The agent-driven framework can interpret abstract descriptors such as "a teacher," "a medical professional," or "an explorer," and translate them into detailed avatar components. This enables generation of diverse avatars that vary in body shape, facial structure, hairstyle, clothing, and accessories. Additional examples illustrating this output diversity are provided in the supplemental materials. Furthermore, as shown in Figure~\ref{fig:poses}, all generated avatars are fully rigged and can be dynamically posed, making them readily usable in animation workflows.

\begin{figure}
    \centering
    \includegraphics[width=1\linewidth]{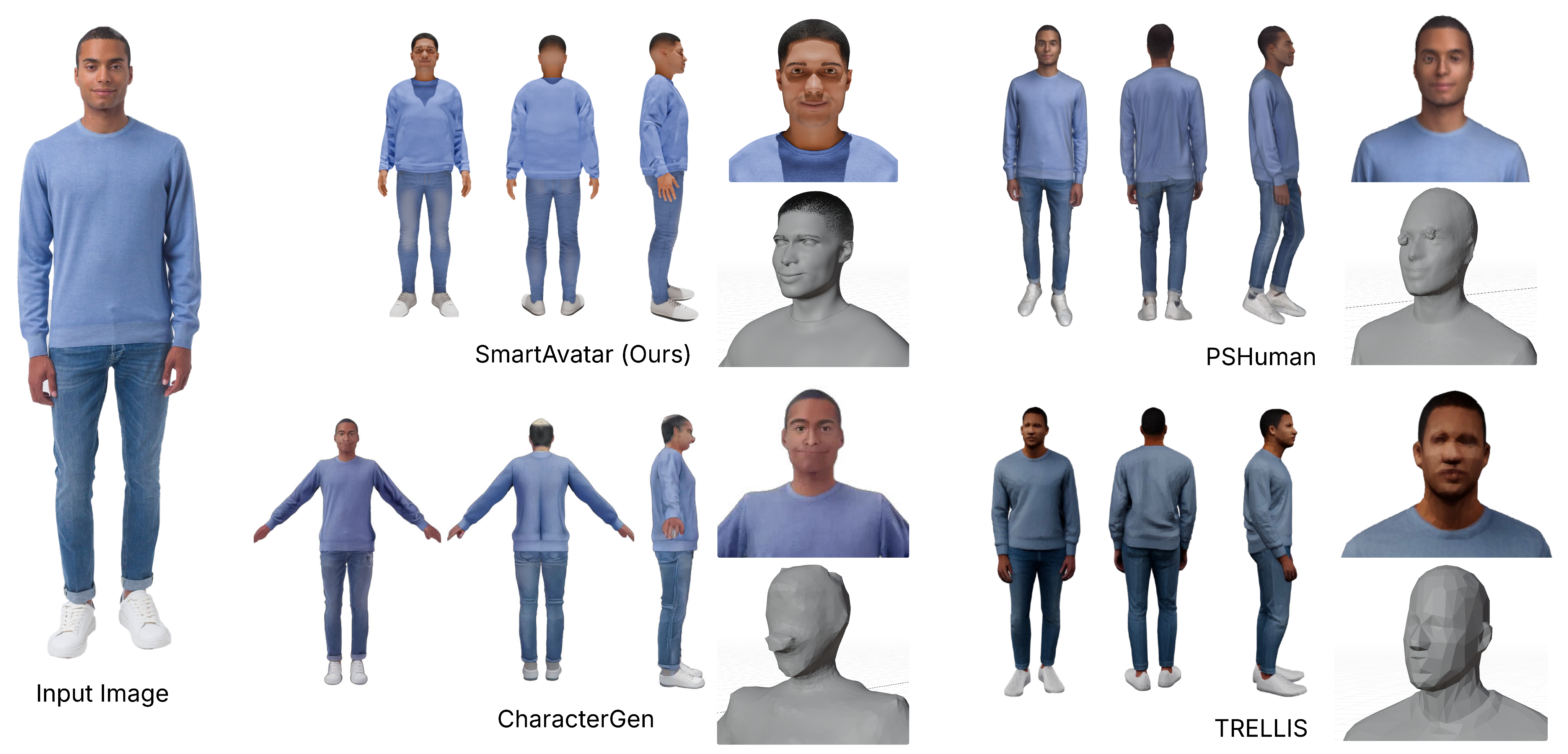 }
    \vspace{-0.3in}
    \caption{Image-to-Avatar Comparison. Our method is compared against PSHuman~\cite{li2025pshumanphotorealisticsingleimage3d}, CharacterGen~\cite{peng2024charactergenefficient3dcharacter}, and TRELLIS~\cite{xiang2025structured3dlatentsscalable}. For better visualization of the reconstructed 3D models, we also show the meshes without textures.}
    \vspace{-0.2in}
    \label{fig:image_input}
\end{figure}

\noindent\textbf{Image-to-Avatar Reconstruction}
SmartAvatar demonstrates strong performance in reconstructing avatars from visual inputs, including portrait and full-body images. As shown in Figure~\ref{fig:image_input}, the system captures salient features such as facial structure, hairstyle, and clothing. The iterative refinement loop enhances fidelity through similarity-guided corrections. For full-body inputs, the auto-verification module incorporates vision-language feedback to refine garment type, color, and style—yielding avatars that closely resemble the source image in both geometry and appearance.

\noindent\textbf{Avatar Editing via Language}
SmartAvatar supports post-generation avatar editing through natural language commands. Figure~\ref{fig:teaser} illustrates this process: users can make adjustments such as “make her surprised” or “make her look younger” These requests are interpreted by the refiner agent, which modifies the Blender code while preserving identity constraints. This supports interactive workflows in character design and customization.

\noindent\textbf{Comparison with State-of-the-Art Methods} 
We compare SmartAvatar with recent state-of-the-art avatar generation models. For text-based input, we evaluate against DreamHuman~\cite{kolotouros2023dreamhuman}, HumanGaussian~\cite{Liu_2024_CVPR}, TADA~\cite{liao2023tadatextanimatabledigital}, and DreamGaussian~\cite{tang2023dreamgaussian}. For image-based input, we compare with CharacterGen~\cite{peng2024charactergenefficient3dcharacter}, PSHuman~\cite{li2025pshumanphotorealisticsingleimage3d}, and TRELLIS~\cite{xiang2025structured3dlatentsscalable}.

As shown in Figure~\ref{fig:image_input}, SmartAvatar generates anatomically coherent avatars with consistent identity features, particularly in the facial region. In contrast, competing models often exhibit mesh distortions such as blurred facial geometry or noticeable outlier bumps across the surface. These visual artifacts are frequently masked by well-aligned textures in forward-facing views, which contributes to higher similarity scores despite the underlying inaccuracies.

SmartAvatar benefits from a modular pipeline that combines procedural reasoning, iterative refinement, and code-based synthesis. This allows for greater structural fidelity and fine-grained control over avatar attributes, including precise adjustments to facial features and body proportions. Unlike end-to-end diffusion-based approaches, our system supports editable, riggable output that can be adapted for downstream tasks.

\begin{figure}[t]
    \centering
    \includegraphics[width=1\linewidth]{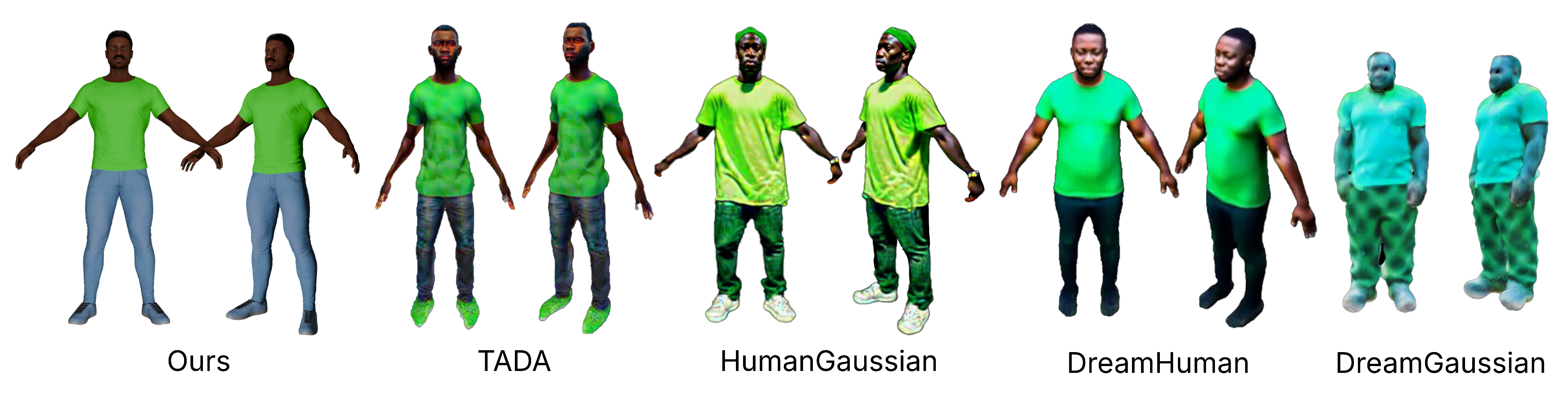}
    \vspace{-0.2in}
    \caption{Text-to-Avatar Comparison. Input Prompt: "A black man wearing a green tee shirt".}
    \label{fig:text_input}
\end{figure}

In text-to-avatar scenarios, illustrated in Figure~\ref{fig:text_input}, other methods generally lack the ability to produce anatomically accurate or customizable avatars. SmartAvatar supports detailed manipulation through natural language commands, enabled by its agent-based refinement loop.

Table~\ref{tab:similarity} reports quantitative metrics including $\text{CLIP}_{\text{text}}$ (similarity between input text and rendered image), $\text{CLIP}_{\text{image}}$ (similarity between input image and rendered image), and ArcFace ID similarity. While diffusion-based models sometimes yield higher numerical scores, this is primarily due to improved texture alignment rather than accurate 3D geometry. Visual inspection confirms that SmartAvatar offers more accurate geometry and higher structural realism.

\begin{table}[t]
  \centering
  \caption{Similarity scores between rendered forward-facing avatars and their corresponding image or text inputs. Although some methods achieve higher embedding metrics, these results are primarily due to texture alignment in frontal views and do not totally reflect the underlying mesh quality, which often contains anatomical distortions and surface artifacts.} 
  \label{tab:similarity}
  \begin{tabular}{l c | l c c}
    \toprule
    \multicolumn{2}{c|}{\textbf{Text Input}} 
    & \multicolumn{3}{c}{\textbf{Image Input}} \\
    \midrule
    Method & CLIP$_{\text{text}}$
      & Method & ArcFace ID Similarity & CLIP$_{\text{image}}$ \\
    \midrule
    Ours          & 0.657   & Ours         & 0.65 & 0.903 \\
    DreamHuman    & 0.657   & PSHuman      & \textbf{0.79}          & 0.918 \\
    HumanGaussian & 0.658   & CharacterGen & 0.66          & \textbf{0.932} \\
    TADA          & \textbf{0.670} & TRELLIS  & 0.50 & 0.887   \\
    DreamGaussian & 0.638  &  &  & \\
    \bottomrule
  \end{tabular}
  \vspace{-0.1in}
\end{table}

    

%% file: sections/ablation_study.tex
\section{Ablation Study}

To quantify the contribution of each component in our prompting pipeline, we perform two ablations: removing chain-of-thought (CoT) reasoning and disabling iterative refinement. Table~\ref{tab:ablation} reports the results in terms of ArcFace ID Similarity and $\text{CLIP}_{image}$.

\begin{table}[ht]
  \centering
  \caption{Component Ablation results. We report mean performance (in \%) for ArcFace ID Similarity and $\text{CLIP}_{image}$ over the test set.}
  \label{tab:ablation}
  \begin{tabular}{lcc}
    \toprule
    \textbf{Condition}            & \textbf{ArcFace ID Similarity} & \textbf{CLIP$_{\text{image}}$} \\
    \midrule
    Full pipeline                 &  \textbf{0.52}                               & \textbf{0.809}                               \\
    \;\;– without CoT reasoning   &  0.48                               & 0.796                               \\
    \;\;– without refinement loop &  0.42                               & 0.772                               \\
    \bottomrule
  \end{tabular}
  \vspace{-0.1in}
\end{table}
\begin{table}[ht]
  \centering
  \caption{Full‐pipeline performance across different vision–language models (VLMs) with reasoning capabilities. }
  \label{tab:vlm-comparison}
  \begin{tabular}{lcc}
    \toprule
    \textbf{Model} & \textbf{ArcFace ID Similarity} & CLIP$_{\text{image}}$ \\
    \midrule
    GPT-4o       & 0.520 & \textbf{0.809}                                \\
    o4-mini       & 0.506 & \textbf{0.809}                                \\
    Gemma 3       & \textbf{0.562} & 0.767                               \\
    \bottomrule
  \end{tabular}
  \vspace{-0.1in}
\end{table}

\noindent\textbf{Chain-of-Thought Ablation.}
We remove all intermediate reasoning steps from the prompt, issuing only the final generation directive. This variant reduces ArcFace ID Similarity by \(7.6\)\% and CLIP Image Similarity by \(1.6\)\% compared to the full pipeline (Table~\ref{tab:ablation}). Qualitatively, we observe increased pattern distortions and degraded identity preservation.

\noindent\textbf{Iterative Refinement Ablation.}
We replace the multi-step feedback loop with a single-shot prompt. This change leads to \(19.2\)\% drop in ArcFace ID Similarity and \(4.7\)\% in CLIP Image Similarity (Table~\ref{tab:ablation}). The resulting models often exhibit coarse geometry and fail to capture fine surface details.

\noindent\textbf{VLM Generalization.}  
To evaluate the generality of our agent-based framework across different vision–language models, we substitute GPT‑4o with o4‑mini and Gemma 3, keeping all other components fixed. We also tested LLaMA 4, but its outputs did not converge to usable reconstructions in our pipeline, indicating that some VLMs may lack the necessary  spatial reasoning and fine-grained control. As shown in Table~\ref{tab:vlm-comparison}, our modular agent-based approach consistently improves performance across compatible models. These results highlight not only the flexibility of our framework, but also its potential as a diagnostic benchmark for probing the spatial and reasoning capabilities of vision–language models in controllable 3D avatar generation.

\noindent\textbf{Summary of Findings.}
Our ablation studies reveal that each component of the SmartAvatar framework contributes a statistically significant improvement to overall performance. Among them, the iterative refinement loop yields the most substantial gains in both identity fidelity and attribute alignment, underscoring the importance of closed-loop reasoning and feedback. Chain-of-thought prompting plays a key role in maintaining semantic and visual consistency, particularly in preserving identity features across views. The structured code template ensures reliable and interpretable synthesis by preserving syntactic correctness in Blender-executable outputs. Furthermore, while the agent-based architecture proves robust across a range of VLM backbones, employing more capable vision–language models leads to noticeably better reconstruction quality. These results collectively highlight the modularity, extensibility, and effectiveness of our framework in high-fidelity, controllable 3D avatar generation.

\noindent\textbf{Limitation}
SmartAvatar builds on an off-the-shelf parametric human generation engine~\cite{humgen3d}, which constrains the realism of the generated avatars. This limits the system's ability to fully capture nuanced facial features, and high-frequency details. Future work could incorporate more expressive human modeling tools or fine-tune the generation pipeline to enhance avatar fidelity and realism.

%% file: sections/conclusion.tex
\section{Conclusion}
We presented \textbf{SmartAvatar}, a modular vision-language-agent framework for generating fully rigged 3D human avatars from either text or image input. The system coordinates a set of specialized VLM/LLM agents, including a semantic descriptor, a code generator, an evaluator, and a refiner, to translate high-level multimodal prompts into Blender-executable scripts. A similarity-based verification loop, combined with chain-of-thought prompting, enables iterative refinement and controllable synthesis of both facial and clothing features. Our qualitative experiments demonstrate SmartAvatar's ability to support diverse inputs, accurate attribute alignment, and flexible post-editing. While the realism of the avatars is bounded by the underlying human generation engine, the modular design provides a strong foundation for future improvements. Ongoing work will focus on enhancing visual fidelity, broadening identity representation, and integrating more expressive modeling backends.

%% file: sections/supp.tex
\section*{Appendix}

\section{Implementation Details}

\subsection{Agent Coordination Pipeline}
Our system coordinates four modular agents (\texttt{Descriptor}, \texttt{Generator}, \texttt{Evaluator}, \texttt{Refiner}) in a verification loop. An additional \texttt{Editor} agent is available for editing the avatar after the initial SmartAvatar pipeline. All agents were instantiated using either GPT-4o or o4-mini. Prompts were engineered to follow a structured chain-of-thought (CoT) format and include few-shot demonstrations retrieved via embedding similarity.

\paragraph{Descriptor Agent.}
The \texttt{Descriptor} converts text, images, or both into a structured attribute representation aligned with the HumGen3D API schema.  
\begin{itemize}
    \item \textbf{Text-Only Inputs:} Uses CoT reasoning to infer body type, attire, and appearance from high-level descriptions.
    \item \textbf{Image-Only Inputs:} Extracts visible features (skin tone, hairstyle, clothing) using multimodal CoT prompting.
    \item \textbf{Multimodal Inputs:} Fuses attributes from both sources with text having override priority when conflicts arise.
    \item \textbf{Attribute Normalization:} Internally maps free-form text to relevant discrete API-compatible tokens (e.g., \texttt{hair\_color="brown"}).
\end{itemize}

\subsection{Generator}

The \texttt{Generator} agent receives a structured attribute representation (e.g., from the \texttt{Descriptor}) and synthesizes Blender-compatible Python code that constructs the corresponding avatar. To improve code reliability and interpretability, the agent is guided by a multi-step chain-of-thought (CoT) prompt that decomposes the task into three phases: (1) reasoning over asset selection and parameter mapping, (2) synthesizing executable code, and (3) validating the code against syntax errors.

\subsection{Evaluator}
The \texttt{Evaluator} agent checks alignment between the rendered avatar and the input prompt or image and decides if the output is similar enough to terminate the refinement loop. We prompt the evaluator agent to analyze the similarity between the rendered portrait and the original input image or text, and output \texttt{\# NO\_CHANGES} if the accuracy score is $\geq \tau$. Otherwise, the loop will continue. The default prompted value for $\tau$ is $90\%$. 

\subsection{Refiner}

The \texttt{Refiner} agent receives the renderings of the current avatar, the corresponding code, and the original input (image or text). It analyzes the differences between the avatar and the original input to identify differences in the set of controllable features.

After identifying the changes that are needed, the \texttt{Refiner} generates updated Blender-compatible Python code to improve the corresponding avatar. Similarly to the \texttt{Generator}, the agent is guided by a multi-step chain-of-thought (CoT) prompt that decomposes the task into three phases: (1) reasoning over asset selection and parameter mapping, (2) synthesizing executable code, and (3) validating the code against syntax errors.

\subsection{Blender and HumGen3D Integration}
We used Blender 4.4 in headless mode with the open-source \texttt{HumGen3D} plugin to generate and manipulate avatars. All avatar synthesis occurred via generated Python code. We provide a shortened snapshot of the code in Figure~\ref{shaded:generate_code}.
\begin{figure}[h!]
    \begin{shaded}
        import bpy
        
        from HumGen3D import Human
        
        \medskip
        
        my\_human = Human.from\_preset("models/female/Hispanic/Tara.json")
    
        \# Customizing body shape
        
        for key in my\_human.body.keys:
        
        \quad if key.name == "Neck Length":
        
        \quad\quad        key.value = 0.5
    
        \quad ...
    
        \# Customizing facial features
        
        for key in my\_human.face.keys:
        
        \quad    if key.name == "eye\_tilt":
            
        \quad\quad        key.value = 0.05
    
        \quad ...
    
        \# Set height
        
        my\_human.height.set(value\_cm=165)
    
        \medskip
        
        \# Set age
        
        my\_human.age.set(32, realtime=False)
    
        \medskip
    
        \# Set hair
        
        my\_human.hair.set\_hair\_quality("high")
        
        my\_human.hair.update\_hair\_shader\_type("accurate")
        
        my\_human.hair.regular\_hair.set("hair/head/female/Long/Bun.json") 
    
        \# Blonde
        
        my\_human.hair.regular\_hair.hue.value            = 0.55
        
        my\_human.hair.regular\_hair.lightness.value      = 3.00
        
        my\_human.hair.regular\_hair.redness.value        = 1.00
        
        \medskip
        
        \# Set skin texture
        
        my\_human.skin.texture.set("textures/female/Default 4K/Female 07.png")  \# Medium-deep warm tone
    
        \medskip
        ...
    \end{shaded}
    \caption{Sample python Blender code for generating an avatar.}
    \label{shaded:generate_code}
\end{figure}

\subsection{Evaluation Metrics}
To evaluate output fidelity, we used ArcFace ID similarity, CLIP image similarity, and CLIP text similarity. Cosine similarity was normalized to $[0,1]$. Facial similarity was calculated from aligned portrait crops unless otherwise noted.

\section{Avatar Diversity}

Figure ~\ref{fig:avatar_diversity} demonstrates the wide variety of phenotypes that can be generated by the human generator and our pipeline. This includes wide ranges of height, age, weight, race, hair, and other features.

\begin{figure}[h!]
    \centering
    \includegraphics[width=1.0\linewidth]{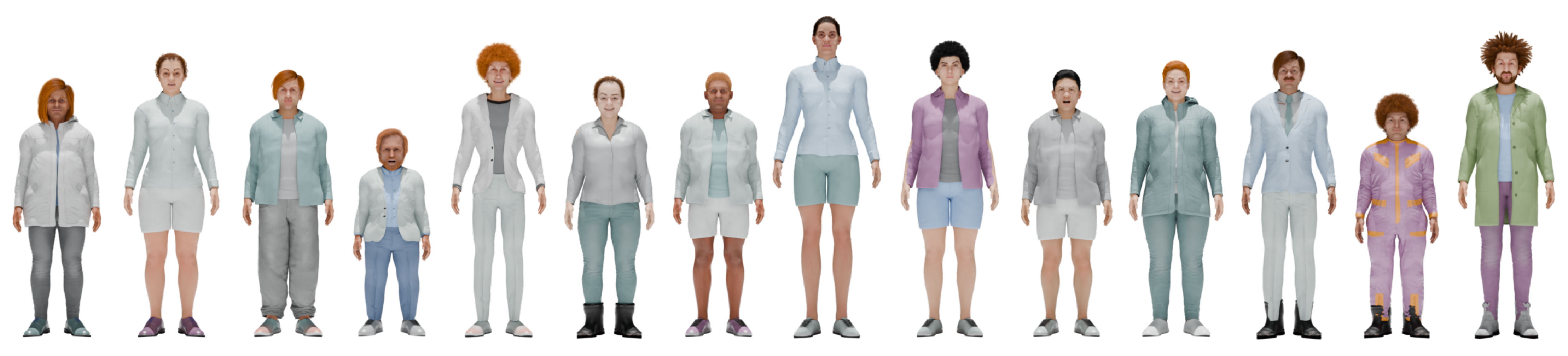}
    \caption{Sample diversity of avatar phenotypes.}
    \label{fig:avatar_diversity}
\end{figure}

As presented earlier, the avatars are rigged and support highly customizable and transferable poses.

\begin{figure}[h!]
    \centering
    \includegraphics[width=1.0\linewidth]{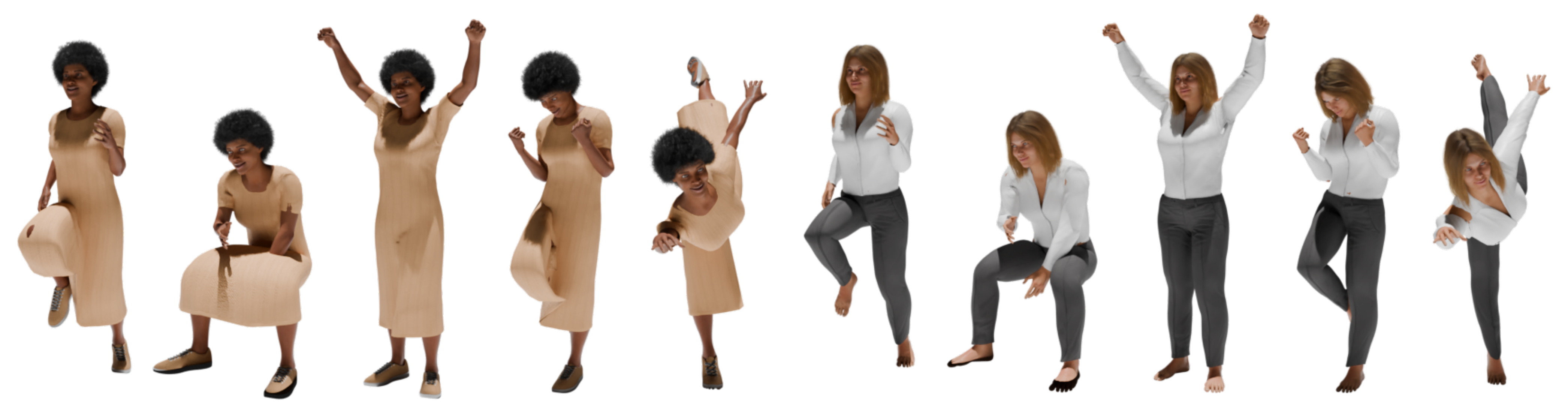}
    \caption{Sample avatar poses transferred across different avatars.}
    \label{fig:avatar_diversity}
\end{figure}

\section{Avatar Generation Examples}

We present qualitative examples to demonstrate SmartAvatar's ability to handle different types of inputs—text-only, image-only, and multimodal combinations—and generate identity-consistent avatars accordingly.

\subsection{Image-Only Inputs}

When given only a portrait or full-body image, SmartAvatar extracts visible appearance features using a vision-language model and generates an identity-consistent 3D avatar. The iterative refinement loop adjusts parameters such as skin tone, hairstyle, and facial structure to better match the reference image.

\begin{figure}[h!]
    \centering
    \includegraphics[width=1.0\linewidth]{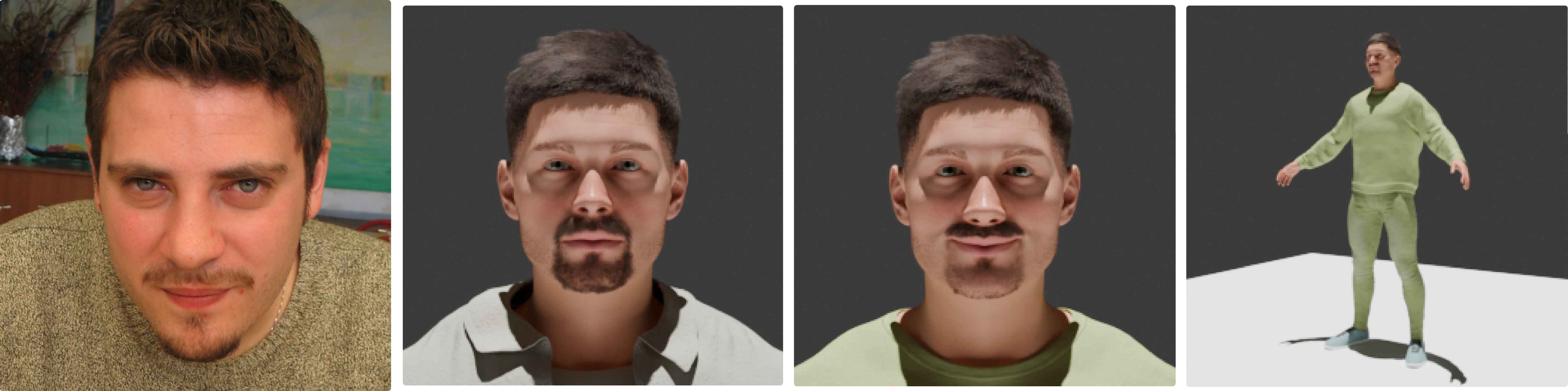}
     \includegraphics[width=1.0\linewidth]{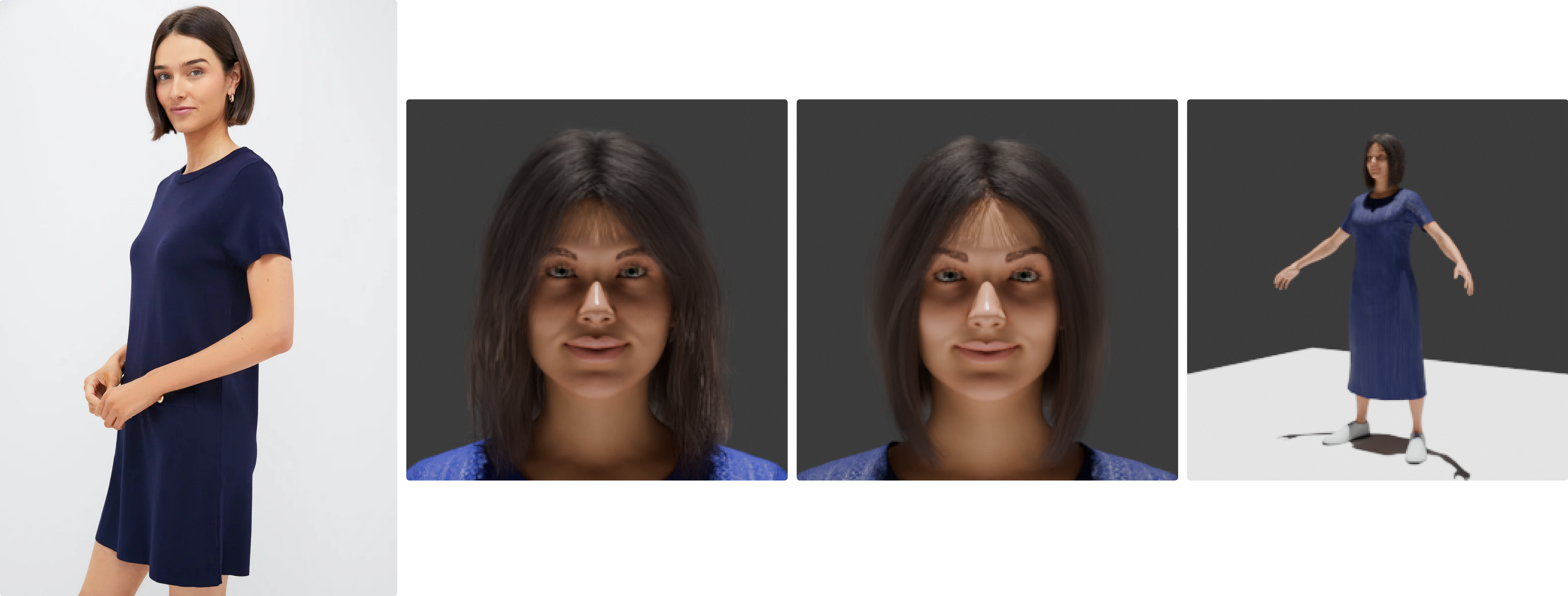}
    \caption{Avatar generation from an image-only input. Each row shows a different subject. Left to right: image input, initial generation, final render after iterative refining.}
    \label{fig:avatar_variation_image}
\end{figure}

\subsection{Text-Only Inputs}

In the absence of an image, SmartAvatar infers attributes from freeform descriptions using chain-of-thought reasoning. This enables controllable avatar synthesis even with abstract or context-rich prompts (e.g., “a middle-aged scientist with gray hair and a lab coat”).

\begin{figure}[h!]
    \centering
    \includegraphics[width=1.0\linewidth]{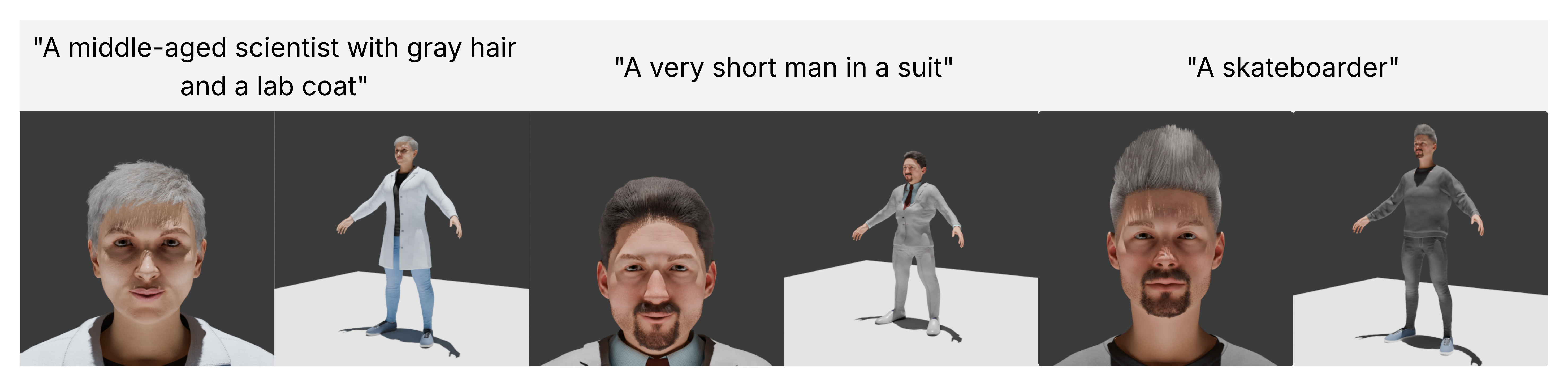}
    \caption{Avatar generation from text-only prompts. Prompts vary in occupation, style, and/or appearance; final avatars reflect semantic differences across descriptions.}
    \label{fig:avatar_variation_text}
\end{figure}

\subsection{Multimodal Inputs}

SmartAvatar can jointly reason over text and image inputs to produce refined avatars. When the text specifies a modification (e.g., “shaved head”), the system resolves conflicts by prioritizing text intent over visual features.

\begin{figure}[h!]
    \centering
    \includegraphics[width=0.75\linewidth]{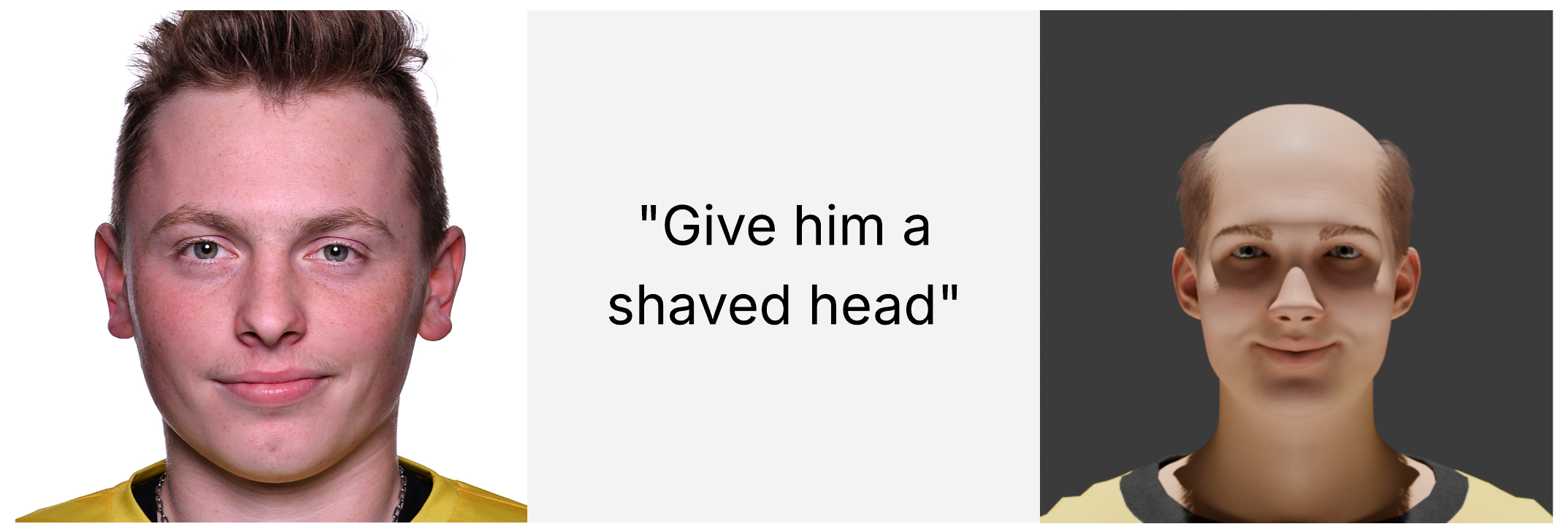}
    \caption{Avatar generation from multimodal inputs. Left: original input image; Middle: text override (“Give him a shaved head”); Right: resulting avatar after fusion and refinement.}
    \label{fig:avatar_variation_multimodal}
\end{figure}

\subsection{Editing Diversity}
\begin{figure}[h!]
    \centering
    \includegraphics[width=1.0\linewidth]{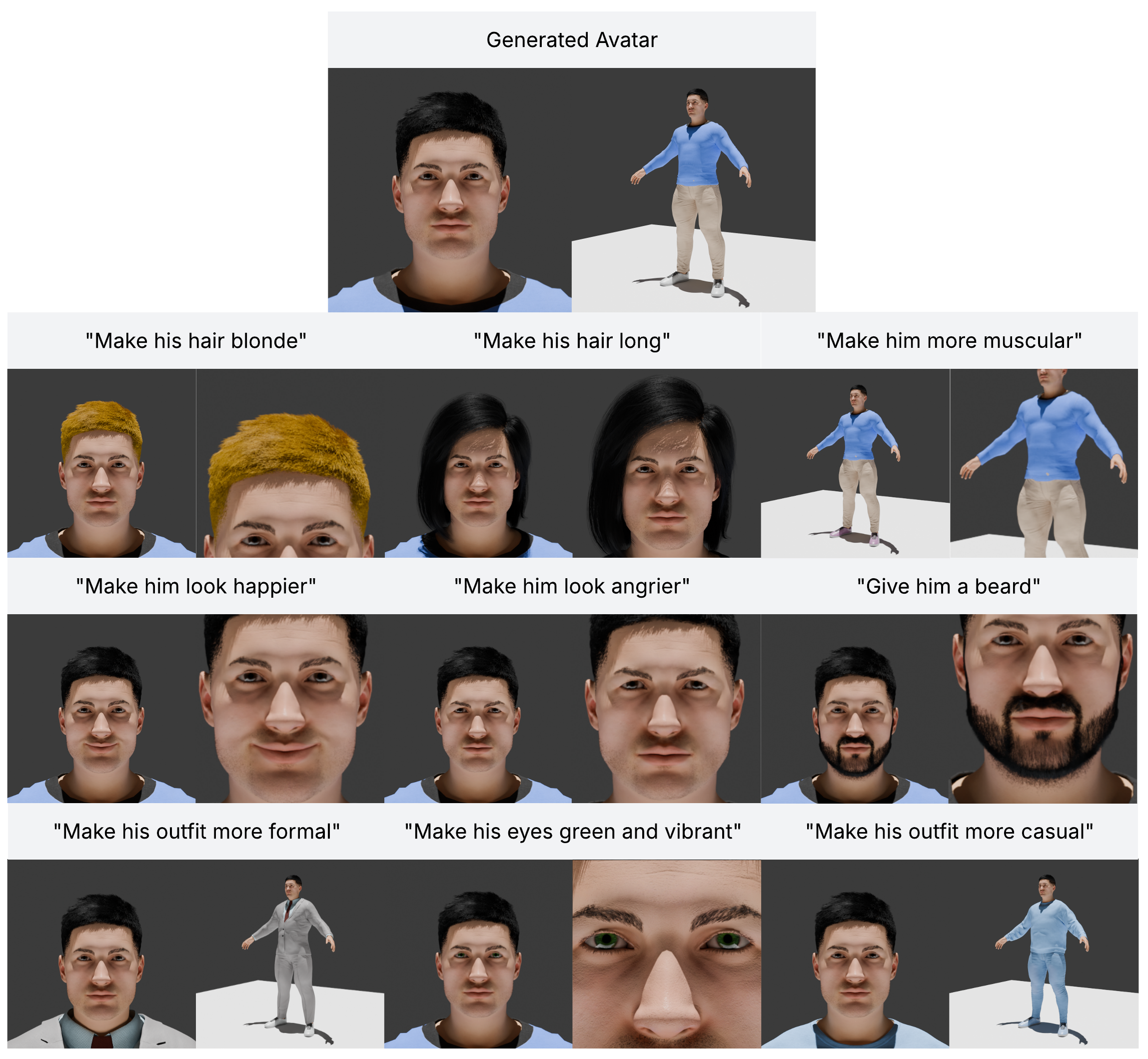}
    \caption{Editing results of a single avatar with diverse editing prompts.}
    \label{fig:editing_diversity}
\end{figure}

Figure~\ref{fig:editing_diversity} shows a single avatar with various editing prompts and the resulting modified avatar. This shows example freeform edits and how they affect rendered outputs, demonstrating conversational control and attribute retention.

